\documentclass[12pt,letterpaper]{article}
\usepackage[a4paper, total={7in, 10in}]{geometry}

\usepackage{graphicx}
\usepackage{helvet}
\usepackage{authblk}
\usepackage{hyperref}
\usepackage{amsmath} 
\usepackage{amssymb} 
\usepackage{orcidlink} 
\usepackage[super,comma,sort&compress]  
   {natbib}
\usepackage[right]{lineno} \linenumbers

\usepackage[utf8]{inputenc} 
\usepackage[T1]{fontenc}    
\usepackage{url}            
\usepackage{booktabs}       
\usepackage{amsfonts}       
\usepackage{nicefrac}       
\usepackage{microtype}      
\usepackage{xcolor}         
\usepackage{multirow}
\usepackage{threeparttable}
\usepackage{amsmath}
\usepackage{subcaption}
\usepackage{makecell}
\usepackage{placeins}

\makeatletter
\renewcommand{\maketitle}{\bgroup\setlength{\parindent}{0pt}
\begin{flushleft}
  \textbf{\@title}
  
  \@author
\end{flushleft}\egroup}
\makeatother

\title{DiffuSETS: 12-lead ECG Generation Conditioned on Clinical Text Reports and Patient-Specific Information}
\date{}

\author[1,2,\#\orcidlink{0009-0007-2648-4938}]{Yongfan Lai}
\author[3,4,\#]{Jiabo Chen}
\author[5]{Deyun Zhang}
\author[5]{Yue Wang}
\author[5]{Shijia Geng}
\author[1,2,*]{Hongyan Li}
\author[3,6,*,\orcidlink{0000-0001-7521-5127}]{Shenda Hong}

\affil[1]{State Key Laboratory of General Artificial Intelligence, Beijing, China}
\affil[2]{School of Intelligence Science and Technology, Peking University, Beijing, China}
\affil[3]{National Institute of Health Data Science, Peking University, Beijing, China}
\affil[4]{VCIP, CS, Nankai University, Tianjin, China}
\affil[5]{HeartVoice Medical Technology, Hefei, China}
\affil[6]{Lead contact}

\affil[$\#$]{These authors contributed equally}
\affil[*]{Correspondence: hongshenda@pku.edu.cn, leehy@pku.edu.cn}

\begin{document}

\maketitle

\section*{SUMMARY}

Heart disease remains a significant threat to human health. As a non-invasive diagnostic tool, the electrocardiogram (ECG) is one of the most widely used methods for cardiac screening. However, the scarcity of high-quality ECG data, driven by privacy concerns and limited medical resources, creates a pressing need for effective ECG signal generation. Existing approaches for generating ECG signals typically rely on small training datasets, lack comprehensive evaluation frameworks, and overlook potential applications beyond data augmentation. To address these challenges, we propose DiffuSETS, a novel framework capable of generating ECG signals with high semantic alignment and fidelity. DiffuSETS accepts various modalities of clinical text reports and patient-specific information as inputs, enabling the creation of clinically meaningful ECG signals. Additionally, to address the lack of standardized evaluation in ECG generation, we introduce a comprehensive benchmarking methodology to assess the effectiveness of generative models in this domain. Our model achieve excellent results in tests, proving its superiority in the task of ECG generation. Furthermore, we showcase its potential to mitigate data scarcity while exploring novel applications in cardiology education and medical knowledge discovery, highlighting the broader impact of our work. 

\section*{KEYWORDS}


Cardiology, Electrocardiogram, Signal processing, ECG generation, Diffusion models

\section*{INTRODUCTION}

\begin{figure}
  \includegraphics[width=1.0\linewidth]{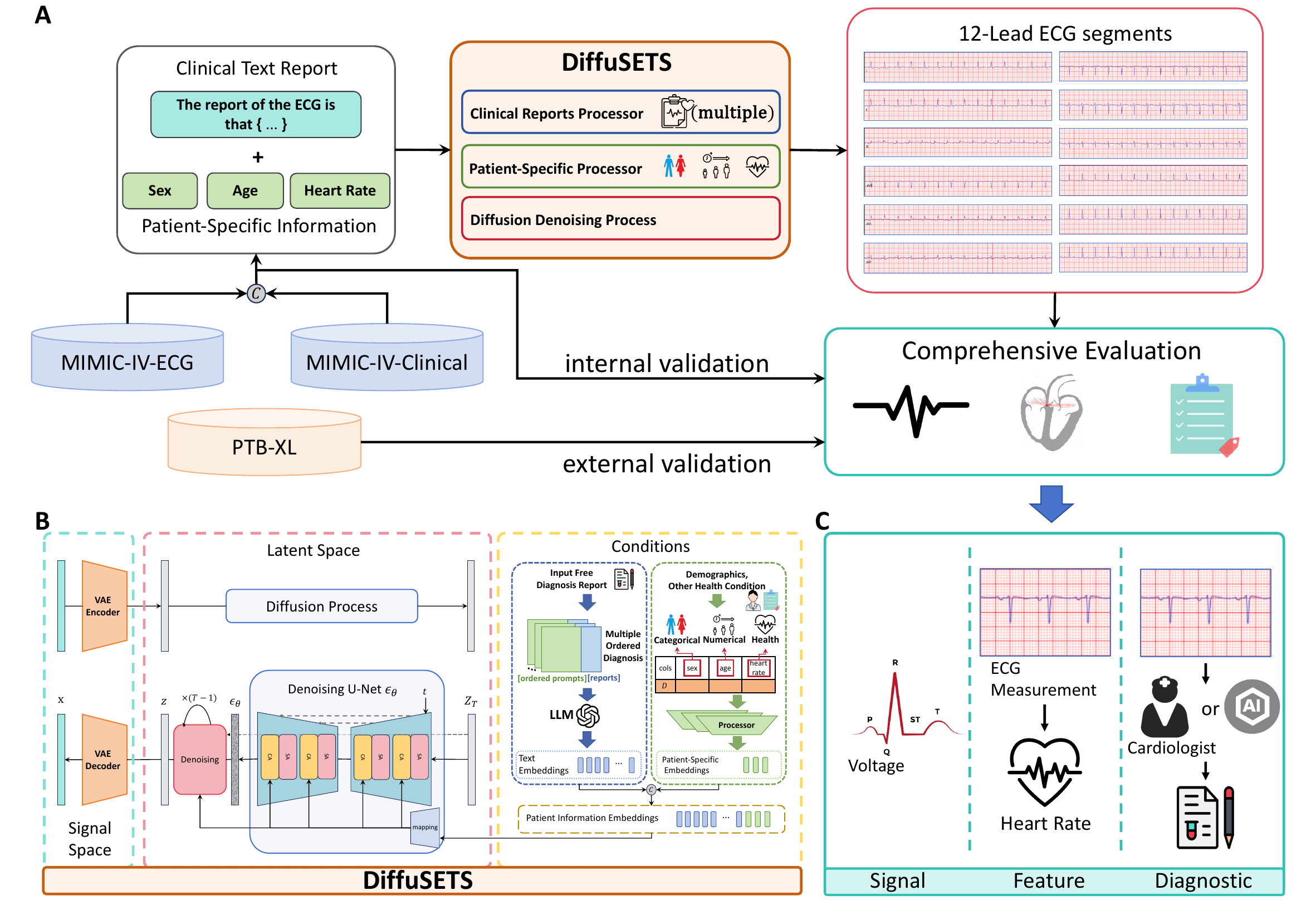}
  \caption{High level overview of our work. (A) We propose DiffuSETS, a 12-lead ECG signal generation model conditioned on clinical text reports and patient-specific information. The model is evaluated through a comprehensive three-level testing process on both internal and external datasets. (B) DiffuSETS model architecture. (C) Three levels of our comprehensive evaluation.}
  \label{fig: 1}
\end{figure}

Currently, cardiovascular diseases account for about one-third of all global deaths \citep{roth2018global}, making them one of the major threats to human health. The electrocardiogram (ECG) is a non-invasive diagnostic tool for heart disease and is widely used in clinical practice \citep{holst1999confident}. This popular assessment provides a comprehensive diagnosis of cardiac activity, but it also places heavy burdens on cardiologists for viewing and interpreting the data. To alleviate this pressure, Many studies have focused on developing ECG classifiers \citep{bian2022identifying, golany202112, kiranyaz2015real} and using them for automated ECG diagnosis. However, acquiring and sharing real ECG signals presents potential risks due to patient privacy concerns\citep{hossain2021ecg, hazra2020synsiggan}. Additionally, accurately labeled ECG signals, especially for rare cardiovascular diseases, are scarce \citep{golany2020simgans}, and their acquisition is both resource-intensive and costly \citep{chen2022me}. These limitations hinder the application of prevailing deep learning methods in advancing cardiological research and practice. Given these challenges, a critical upstream task is the generation of synthetic ECG signals \citep{golany2020simgans, zhang2021synthesis}.

In the field of ECG signal generation, the main research goal is to generate ECG signal samples with high fidelity and rich diversity. Many studies have adopted the Generative Adversarial Network (GAN) architecture to generate ECG signals \citep{adib2023synthetic}, and others have introduced Ordinary Differential Equation (ODE) systems representing cardiac dynamics to create ECG samples \citep{golany2020simgans}. Recently, some new studies have incorporated patient-specific cardiac disease information into the ECG generation process to improve the generated outcomes \citep{chen2022me, alcaraz2023diffusion}, and some have included the content of patients' clinical text reports during generation \citep{chung2023text}. However, current research in this field still has certain deficiencies: (1) Dataset limitations. Most of the previous ECG datasets contain a limited variety of samples and have low textual richness in clinical text reports. Therefore, generative models trained on such datasets have limited performance and are unable to generate extremely rare ECG signals that are absent from the training data. Although extremely rare ECG samples remain scarce, the newly released MIMIC-IV-ECG \citep{gowmimic} and MIMIC-IV-Clinical \citep{johnson2023mimic} provide a large ECG dataset with ample related information, paving the way for generation models with more possibilities. (2) Difficulty in unifying features across different modalities. Clinical text reports and patient-specific information encompass data from various modalities, each with distinct distributions, making it challenging to integrate all this information into a model. Consequently, many ECG generation models rely solely on high-level label information, which limits the diversity of the generated outputs. (3) Lack of benchmarking methods. In the field of ECG generation, there is still a lack of comprehensive benchmarking methods, making it difficult to assess the relative merits of models.

Equally significant is the growing interest in expanding the applications of generative models beyond merely providing data for other AI models. The capabilities of generative methods, such as Denoising Diffusion Probabilistic Models (DDPM) \citep{ho2020denoising}, have been demonstrated, and their potential applications in scientific fields like neuroscience are beginning to be explored \citep{generating_eeg}. However, the full potential of generative models in cardiology remains largely untapped. This raises a compelling question: what additional contributions can generative models make to the field of cardiology? Answering this question presents a meaningful challenge that warrants exploration.

To address the aforementioned issues, this paper introduces DiffuSETS, a \textbf{Diffu}sion model to \textbf{S}ynthesize 12-lead \textbf{E}CGs conditioned on clinical \textbf{T}ext reports and patient-\textbf{S}pecific information. Our approach is the first work to use diffusion models to handle ECGs generation from text. DiffuSETS utilizes the MIMIC-IV-ECG dataset as the training dataset, which features a wide variety of characteristics suitable for ECG signal generation and enhances the diversity of the generated signal samples. We also design a comprehensive evaluation, which includes quantitative and qualitative analyses at the signal level, feature level, and diagnostic level. Such testing allows for a comprehensive evaluation of the performance of generative models. We have also incorporated a clinical Turing test, involving evaluations by cardiologists, to ensure the high fidelity of the generated ECG samples. High level overview of our work are shown in Figure~\ref{fig: 1}A. Furthermore, to explore the potential contributions of DiffuSETS, we not only demonstrate its ability to enhance downstream deep learning models but also use it to generate ECG signals under complex cardiac conditions and extremely rare heart diseases for cardiological education. Finally, we rigorously show that DiffuSETS can capture latent causal links between ECG signals and non-cardiac conditions, positioning it as a valuable tool for medical knowledge discovery.

The main contributions of this paper are as follows:  (A) We introduce DiffuSETS, an ECG signal generator that accepts clinical text reports and patient-specific information as inputs. This is the first work to use diffusion models trained on MIMIC-IV-ECG dataset for generating ECGs from text. ECG signal samples with high semantic alignment can be generated by just inputting simple natural language text as a description of the patient's disease information. We can also accept inputs such as heart rate, sex, and age, adding constraints to the features of the generated ECG signals, thus making the generation of ECG signals more detailed and diverse. (B) We have designed a set of comprehensive evaluation to evaluate the effectiveness of ECG signal generation, which can comprehensively assess the performance of generative models. Our method is tested within this comprehensive evaluation and a Turing test. Both results are very significant, demonstrating the fidelity and semantic alignment of the model-generated ECG signal samples. (C) We address the challenge of data scarcity by applying DiffuSETS-generated ECGs to downstream tasks, showing through careful experimentation that the model offers significant benefits to ECG auto diagnosis. Furthermore, we explore the potential of DiffuSETS in promoting cardiological education and facilitating medical discoveries. This endeavor underscores the versatility of our model and the broader significance of our work.

\section*{RESULTS}

\subsection*{DifuSETS overview and Data curation}

In this paper, we propose DiffuSETS, a diffusion model capable of generating 12-lead ECGs from clinical text reports and patient-specific information. 
The architecture of DiffuSETS is illustrated in the Figure~\ref{fig: 1}B, involving three modalities: signal space, latent space, and conditional information space (clinical text reports and patient-specific information). 
The variational autoencoder facilitates the conversion of vectors between signal space and latent space, while a prompted large language model, serving as the semantic embedding model, extracts embedding vectors from clinical text reports to capture semantic information, which are then merged with a embedding vector generated from patient-specific information. Finally, the denoising diffusion process and noise prediction model collaboratively achieve the cross-modal generation from text to electrocardiographic latent variables.

We use the MIMIC-IV-ECG dataset \citep{gowmimic} for training DiffuSETS, which contains $794,372$ 12-lead ECG signal records after preprocessing. From this dataset, we extract heart rate features and corresponding clinical text reports. Then, using patient ID as key, we query the MIMIC-IV-CLINICAL dataset patient table to retrieve patient-specific information, such as sex and age. Additionally, we utilize PTB-XL dataset, which contains $299,712$ samples, for external validation and comparison with other baseline models.

\subsection*{DiffuSETS can generate high fidelity ECG}
\label{evaluation_result}

Our conprehensive evaluation involves experiments and analysis at the signal level, feature level, and diagnostic level. The overall design is shown in Figure \ref{fig: 1}C. In this framework, we aim to ask three key questions regarding ECG signals generated from clinical text reports and patient-specific information:

\begin{enumerate}
    \item (Signal level) Does the generated ECG signal resemble real ECG signals? 
    \item (Feature level) Are the input features accurately reflected in the generated signals? 
    \item (Diagnostic level) How closely does the generated ECG align with the corresponding clinical text report?
\end{enumerate}

These questions guide our evaluation of the effectiveness and relevance of the generated ECG signals in capturing both the clinical context and the underlying physiological characteristics. DiffuSETS acheives remarkable scores in this three level comprehensive evaluation, results are listed in Table~\ref{tab: all_result}.

\begin{figure}
  \centering
  \includegraphics[width=0.93\linewidth]{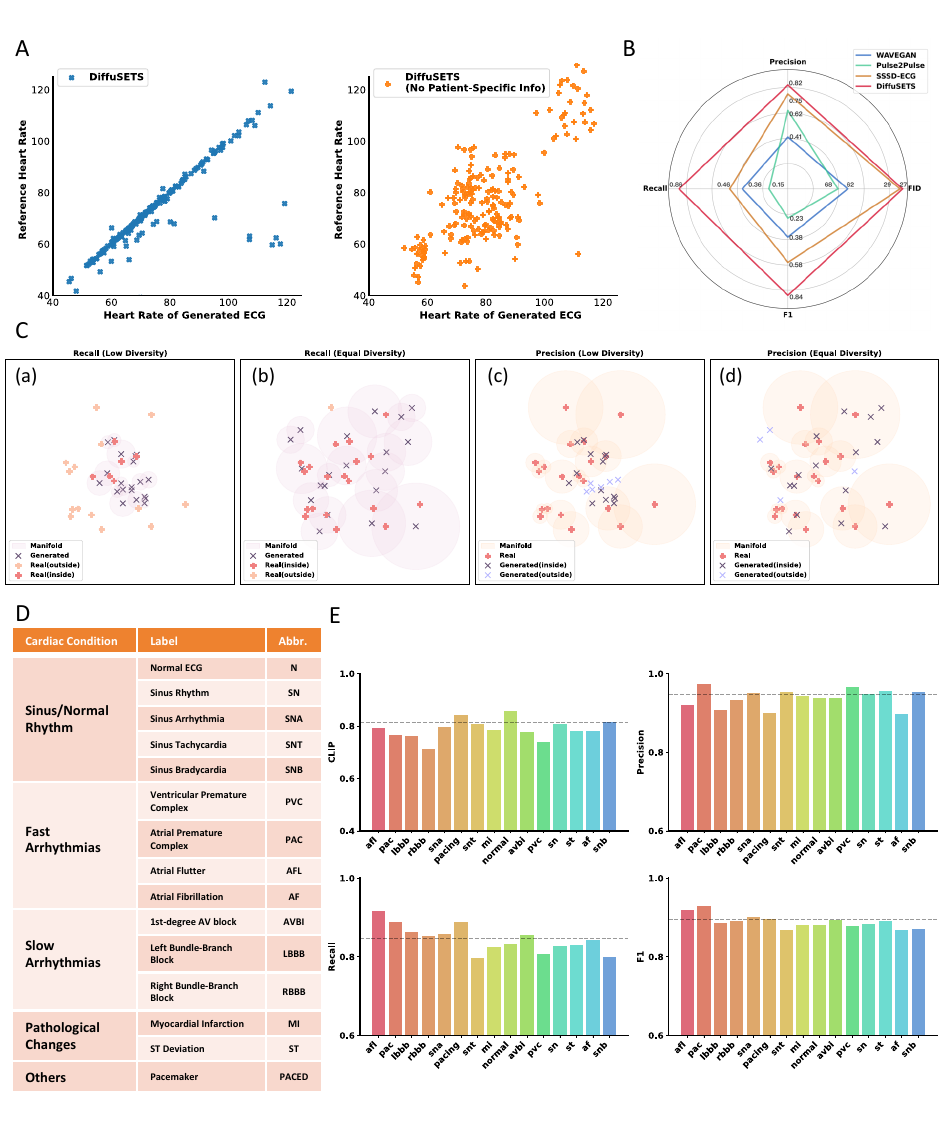}
  \caption{DiffuSETS can generate ECG of high quality and variety. (A) The heart rate scatters of DiffuSETS (left) and DiffuSETS without patient-specific information input (right). (B) The signal level performance of DiffuSETS and other baseline models on PTB-XL. (C) Conceptual illustrations of precision and recall calculations: (a, c) The diversity of the generated results is low, meaning the real signals rarely fall into the manifold constructed by the generated signals, leading to a significant gap between recall and precision. (b, d) The diversity of the generated results matches that of the real signals, resulting in good and balanced precision and recall. (D) The descriptions of chosen labels and related cardiac condition category. (E) Assessments of DiffSETS-generated ECG signals under curated labels. Average performcane are marked in grey dash lines.} 
  \label{fig: 2}
\end{figure}

At the signal level, we focus on the fidelity and stability of the generated signals by evaluating various metrics that assess both the distribution similarity and the structural resemblance between real and generated ECG signals. First, we calculate the Fréchet Inception Distance (FID) score to measure the distribution similarity between the two sets of signals. Since the FID score only assesses data distribution through macro scope of mean and variance, we also evaluate waveform-level similarity to ensure the quality of the generated ECG signals. For phase deviations may occur between the generated and real signals, traditional Mean Absolute Error (MAE) is not a reliable metric for assessing similarity. Instead, we adopt the brilliantly proposed method \cite{kynkaanniemi2019improved}, which calculates precision and recall by counting the ratio of generated ECG signal representation points that fall within the manifold constructed by the real ECG signal representation points, and vice versa (illustrative cases are depicted in Figure~\ref{fig: 2}C). Finally the F1 score is derived to assess the overall performance of accuracy and diversity. We use a pretrained Net1D\cite{hong2020holmes} as the encoder to obtain ECG representations in signal level assessment. The results shows that the ECG signals generated by DiffuSETS resemble the real ECG signals in both signal distribution and representation manifold, which proves the fidelity and stability of generated signals.

\begin{table}[!htbp]
\centering
\caption{Comprehensive evaluation results and comparisons with ablation models. The best values are bolded while the seconded best are underlined.}
\label{tab: all_result}
\resizebox{\columnwidth}{!}{%
\begin{tabular}{c|cccc|c|c}
\toprule
 &\multicolumn{4}{c|}{\textbf{Signal Level}} & \textbf{Feature Level} & \textbf{Diagnostic Level}\\
\textbf{Model} & \textbf{FID (↓)} & \textbf{Precision (↑)} & \textbf{Recall (↑)} & \textbf{F1 Score (↑)} & \textbf{Heart Rate MAE (↓)} & \textbf{CLIP (↑)} \\
\midrule
\multicolumn{7}{c}{\textbf{MIMIC-IV-ECG (internal)}}\\
\midrule
DiffuSETS & \textbf{23.6} & \textbf{0.947} & \textbf{0.847} & \textbf{0.894} & \textbf{4.15} & \underline{0.812} \\
w/o PS-info & 51.1 & \underline{0.942} & \underline{0.841} & \underline{0.889} & 12.25 & 0.799 \\
w/o VAE & \underline{48.9} & 0.894 & 0.661 & 0.759 & \underline{7.04} & \textbf{0.84} \\
w/o PS-info\&VAE & 93.8 & 0.893 & 0.614 & 0.728 & 13.29 & 0.811 \\
\midrule
\multicolumn{7}{c}{\textbf{PTB-XL (external)}}\\
\midrule
DiffuSETS & \textbf{27.6} & \textbf{0.821} & \underline{0.861} & \textbf{0.841} & \textbf{6.73} & \underline{0.795} \\
w/o PS-info & 41.2 & \underline{0.756} & \textbf{0.881} & \underline{0.814} & 13.29 & 0.788 \\
w/o VAE & \underline{32.5} & 0.734 & 0.598 & 0.659 & \underline{9.53} & \textbf{0.83} \\
w/o PS-info\&VAE & 77.5 & 0.67 & 0.536 & 0.596 & 13.52 & 0.776 \\
\bottomrule
\end{tabular}
}
\end{table}

At the feature level, we examine whether the ECG signals generated by the model align with the input descriptions of patient-specific information. Considering that sex and age are difficult to measure through quantitative analysis, we choose heart rate as the focal point for testing. Our approach involves using the condition in ground truth sample as the input and obtaining heart rate value of generated ECG signal. Besides computing the MAE between input heart rate and heart rate of generated ECG signal, we depict scatters plots of the heart rate pairs (Figure~\ref{fig: 2}A) respectively to visualize the result. It is suggested that patient-specific information significantly reduces the heart rate deviation, which demonstrates that our model can generate conditional ECG signals finely based on the heart rate information contained in the patient-specific information. Moreover, the inclusion of patient-specific information also empowers DiffuSETS to perform more flexible generation concerning with physiological conditions, which enlightens the medical discoveries.

At the diagnostic level, we assess whether the generated ECGs conform to the descriptions of the disease, that is, the content of the clinical text reports. We use CLIP scores \citep{clipscore} to assess the semantic alignment between the ECG signals generated by the DiffuSETS and the input clinical text report. This process can be viewed as use a text-ECG encoder that has already achieved semantic alignment in real dataset to measure whether the ECG signals generated by our model match the disease conditions. Similar to the method in signal level assessment, we use the Net1D\citep{hong2020holmes} as ECG encoder and deploy an MLP to project text embedding output by Large language model into the aligned space. In diagnostic assessment, DiffuSETS achieves a substantial performance in clinical alignment, which proves that it learns the hidden relationship between clinical text report and ECG waveform. Additionally, readers are encouraged to revisit the scatter plot in feature level (Figure~\ref{fig: 2}A) and observe the results of the DiffuSETS model without patient-specific information (represented by orange filled plus). Notably, these points tend to form three distinct clusters centered around $(55, 55)$, $(80, 80)$, and $(110, 110)$, despite the absence of heart rate input. This phenomenon suggests that model has learned the rhythmic information from clinical text report, because the clustering in fact corresponds to typical heart rates associated with rhythmic report “sinus bradycardia” (heart rate below normal), “sinus rhythm” (normal heart rate), and “sinus tachycardia” (heart rate above normal).

To evaluate signal quality in greater detail, we utilize DiffuSETS to generate ECG signals corresponding to various cardiac conditions. Since the MIMIC IV ECG dataset lacks class-level labels, we parse the clinical text reports and employ keyword matching to select appropriate samples. The curated labels cover a wide range of common cardiac conditions, as listed in Figure~\ref{fig: 2}D, offering a comprehensive view to test generation and alignment ability of DiffuSETS. For the generated ECG signals, we assess precision, recall, F1 score, and CLIP score, as illustrated in Figure~\ref{fig: 2}E. The results indicate that DiffuSETS produces samples of consistently high quality across most cases. Additionally, the findings suggest that our approach effectively resists data imbalances in the training dataset, as the number of ECG signals varies significantly across different conditions. A plausible explanation is that the model learns fundamental waveform patterns from a large number of training samples, thus only requiring relatively fewer examples to capture distinctive features and adjust outputs to align with specific distributions.

Finally, for external validations, we directly perform the three level test using PTB-XL dataset without modifying models. The result is shown in Table~\ref{tab: all_result}. DiffuSETS can still generate high quality ECG on other clinical text report system without any fine-tuning, which reflects the generalization ability credited to the usage of LLM and our prompt design. 

\subsection*{DiffuSETS outperforms other ECG generation methods}

In addition to comparing DiffuSETS with its ablation models, we conducted an experiment on the PTB-XL dataset to compare it against popular ECG/time series generation methods. While other open-source models cannot generate ECG signals based on clinical text reports or patient-specific information, we evaluated the overall quality of the generated ECG signals using signal level tests. Among the baseline models, all of which condtioned on labels, WAVEGAN\citep{wavegan} is for general time series synthesis, Pulse2Pulse\citep{thambawita2021deepfake} and SSSD-ECG\citep{adib2023synthetic} are models specifically tailored for ECG generation. 

As shown in Figure~\ref{fig: 2}B, our method achieves the highest quality ECG generation in all four metrics. Furthermore, it is worth noting that other baseline models exhibit a noticeable gap between precision and recall scores. We conclude it to the fact that the manifold of ECG signals generated by other methods is much smaller than that of real ECG signals\citep{kynkaanniemi2019improved}, as illustrated in Figure~\ref{fig: 2}C(a)(c). That is, the diversity of the generated ECG signal does not match the real one. However, DiffuSETS does not experience a decline in recall scores, as illustrated in Figure~\ref{fig: 2}C(b)(d), demonstrating the effectiveness of conditioning on clinical text reports and patient-specific information. Moreover, in practical applications, the diversity of DiffuSETS-generated ECGs can be further improved by adjusting the input clinical text reports and patient-specific details — an advantage that models conditioned solely on labels cannot provide.

\subsection*{DiffuSETS has the potential to pass the Turing test}

\begin{figure}
  \centering
  \includegraphics[width=0.8\columnwidth]{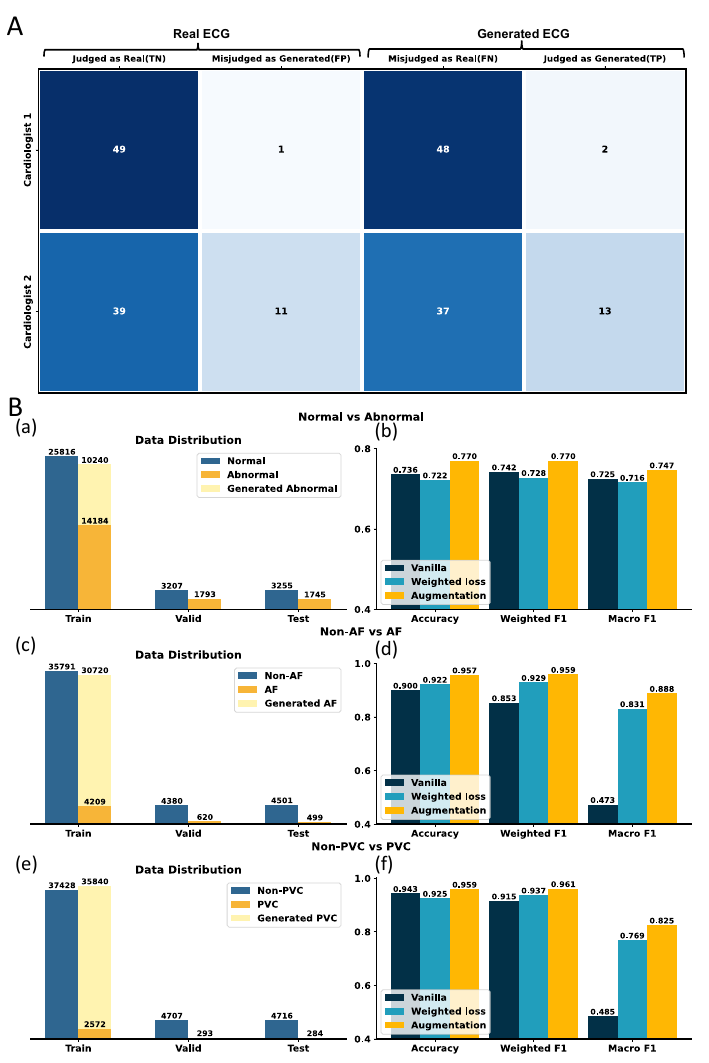}
  \caption{DiffuSETS can pass the turing test and address downstream data scarcity. (A) Judgements for real and generated ECGs of two cardiologists. (B) The data distribution and performance of designed three downstream imbalanced classification task: (a, b) Normal/Abnormal classification. (c, d) AF/Non-AF classification. (e, f) PVC/Non-PVC classification.}
  \label{fig: 3}
\end{figure}

We conduct two kinds of cardiologist evaluation test. In the fidelity evaluation, we extract feature information from $50$ records in the MIMIC-IV-ECG dataset, using as input to generate $50$ ECG signals, and then randomly select another $50$ ECG signals from the MIMIC-IV-ECG dataset. We provide these generated data alongside the real data to cardiologists for Turing test assessment. The cardiologists are tasked with determining whether the provided ECGs are generated by a machine. The judgements are recorded in Figure~\ref{fig: 3}A. In the test for semantic alignment, we provide $100$ generated ECGs (using different conditions, especially clinical text reports, recorded in MIMIC-IV-ECG). Experts are asked to determine whether our generated results matched all the descriptions in the clinical text reports. 

In Turing test, despite different cardiologists may have different criteria for distinguishing real and generated ECG, which results in quantity gap of total number of ECG judged to be model-generated, judgments made by the same cardiologist show non-discrimination for real and generated ECG, i.e. the TP(Generated ECG judged as generated) value is close to FP(Real ECG misjudged as generated) value within the same row. This consistency indicates that even experienced cardiologists struggle to differentiate between real and generated ECG signals. Moreover, the significant proportion of false negatives (generated ECG misjudged as real) further underscores the high fidelity of the ECG signals produced by the DiffuSETS model, making them virtually indistinguishable from real ones. For alignment test, according to expert evaluation, the accuracy of semantic alignment in the electrocardiogram generated by DiffuSETS has reached 76\%. Actually it is a relatively hard task since all reports need to check the consistency before the ECG data can be judged as alignment, there are still more than three-quarter generated ECGs are considered matching to input clinical text reports, which indicates that DiffuSETS model do grasp the diagnostic information between signal and text modalities when expanding the diversity of generated ECG signals. 

\subsection*{DiffuSETS can enhance the performance of ECG diagnosis, particularly for rare diseases}

A key objective of our efforts is to solve the deficiency of labeled ECG signals, particularly for rare cardiovascular diseases. To prove that DiffuSETS can generate medical data with high fidelity and without privacy issues, thus enhancing the potential of deep learning models in downstream tasks, we design three binary classification experiments. For each experiment, we filter out 50,000 ECG samples from the MIMIC-IV-ECG dataset by detecting kerword in the associated clinical text report and then partition to percentage of 80\%, 10\% and 10\% for training, validation and test. Note that all three tasks suffer from varying degrees of data imbalance, with the most severe case exhibiting a quantity ratio of nearly 15:1, reflecting the challenges commonly encountered in real-world deep learning model training. We use DiffuSETS to generate ECG samples for the minority class, thereby creating a more balanced training dataset. Figure~\ref{fig: 3}B(a)(c)(e) shows the statistics of three datasets before and after augmentation. In addition to training directly with the original dataset, we also evaluate the weighted cross-entropy loss method, where the weight for the minority class is set according to the imbalance ratio. We adopt the NET1D\citep{hong2020holmes} as the classifier, with modifications made only to the output layer.

As shown in Figure~\ref{fig: 3}B(b)(d)(f), when trained directly on the original dataset, the model tends to classify all samples into the majority class due to severe data imbalance (evident in the AF/Non-AF and PVC/Non-PVC tasks). The weighted loss method mitigates this issue to some extent but degrades performance when the imbalance is less severe (as seen in the Normal/Abnormal task). In contrast, the model trained on the augmented dataset achieves the best results across all three experiments, effectively addressing the imbalance problem. These findings not only demonstrate the fidelity of the signals generated by DiffuSETS again, but also highlight its potential to enhance downstream tasks, which is one of the key motivations.

\subsection*{DiffuSETS can promote cardiological education and initiate new medical discoveries}

\begin{figure}
  \includegraphics[width=1.0\linewidth]{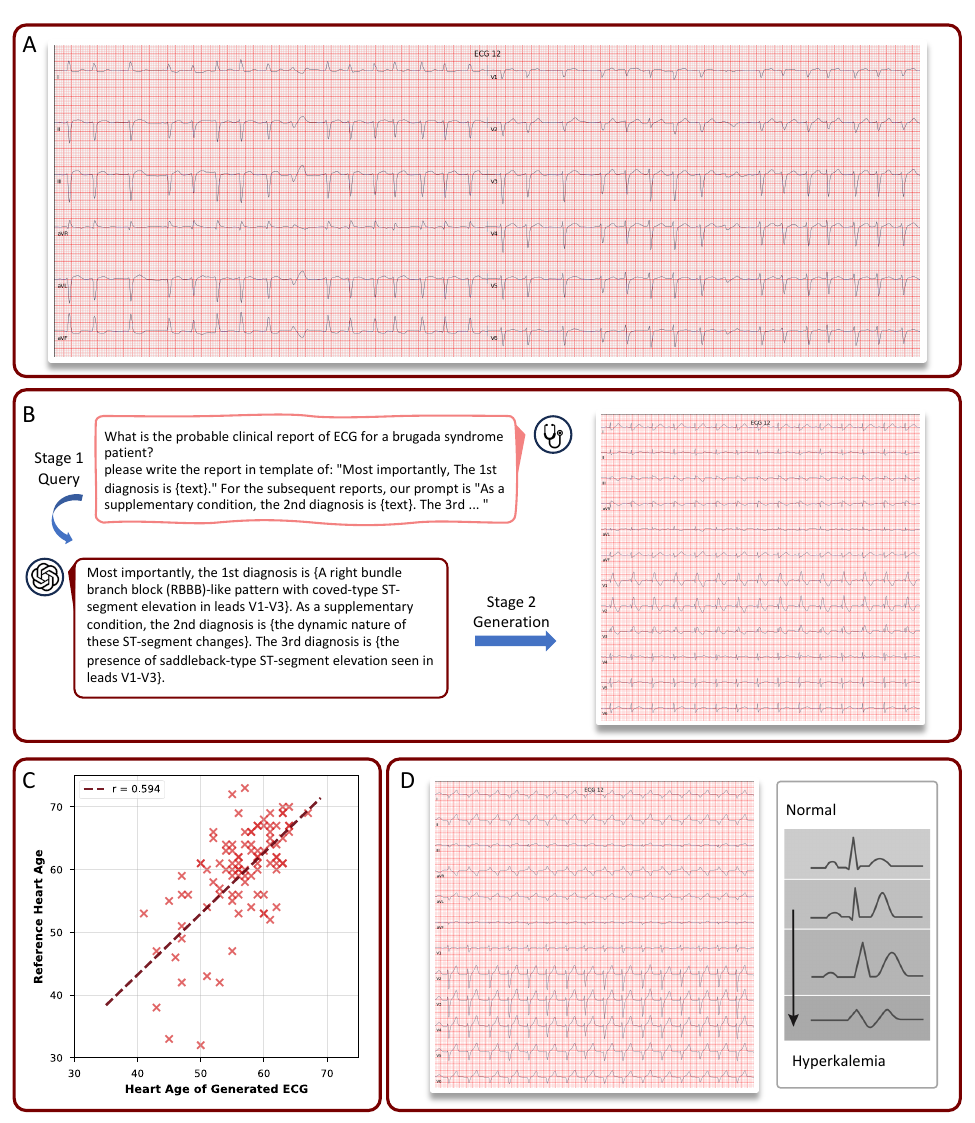}
  \caption{DiffuSETS can promote cardiological education and initiate new medical discoveries. (A) Example of DiffuSETS-generated ECG signal under complex cardiac conditions. (B) The two-stage framework of utilizing DiffuSETS and LLM to generate extremely rare Brugada syndrome. (C) The heart age scatters and correlation assessment. (D) Example of DiffuSETS-generated ECG signal under non-cardiac condition "hyperkalemia"(left) and the documented ECG waveform trends(right) when blood potassium level rises.}
  \label{fig: 4}
\end{figure}

Accurately interpreting ECGs under various heart conditions is a critical component of cardiology education. With the support of DiffuSETS, learners can generate ECG signals tailored to specific conditions of interest, thereby broadening their exposure and understanding. To demonstrate DiffuSETS’ capability to handle complex cardiac conditions, we present a case study of generated ECG signals (Figure~\ref{fig: 4}A) and provide an explanation of how they align with the input clinical text reports. 

The input clinical text report for this electrocardiogram (Figure~\ref{fig: 4}A) is "1. Atrial fibrillation with PVC(s) or aberrant ventricular conduction. 2. Left anterior fascicular block. 3. Anteroseptal infarct - age undetermined. 4. Left ventricular hypertrophy. 5. Lateral ST-T changes may be due to hypertrophy and/or ischemia. 6. Abnormal ECG.", with a heart rate of 79 beats per minute and the male patient is 96 years old. The patient suffers from multiple concurrent severe arrhythmias, making it difficult to generate. Overall, it is an abnormal ECG. The generated ECG has a heart rate of approximately 80 beats per minute, which is consistent with the input information. The patient's RR intervals are irregular, and the P waves are absent, which is characteristic of atrial fibrillation. The QRS complex of the 9th heartbeat is significantly different from the others, being wide and malformed, indicating the occurrence of a ventricular premature beat. Moreover, the overall width of the QRS complex is wider, suggesting that the patient's ventricular conduction is abnormal, specifically diagnosed as left anterior fascicular block. In addition, the patient exhibits changes in the ST segment and T wave, indicating that the patient may have severe symptoms of myocardial ischemia or infarction. Finally, the voltage of the QRS complex is high, suggesting the possibility of left ventricular hypertrophy. This case sufficiently proves that our model is competent to generate ECG under a complex cardiac condition. 

The other significant limitation in current cardiology education is the scarcity of ECG examples for extremely rare heart diseases, such as Brugada syndrome \citep{brugada} and Long QT syndrome \citep{longqt}, both of which are often associated with a high risk of sudden cardiac death. Diagnosing these rare conditions typically relies on memorizing a few waveform patterns outlined in textbooks. Thus, it would be highly beneficial for cardiologists to have access to an ample supply of ECG samples for specific cases. However, the rarity of these diseases poses a challenge for conventional conditional generative learning methods, which require a large number of labeled training samples to model the relationship between conditions and data distributions effectively. To address this issue, we propose a novel framework leveraging DiffuSETS in conjunction with large language models, enabling the generation of ECG signals for rare cardiac conditions (Figure~\ref{fig: 4}B).

The proposed framework operates in two stages. First, a large language model is queried to extract and summarize the ECG abnormalities and clinical characteristics associated with the target condition. Considering the potential hallucination of large language model, if necessary, expert review mechanisms can also be incorporated to ensure correctness. Second, the information is integrated with our custom-designed ordered prompt to generate a text embedding, which DiffuSETS then utilizes, along with predefined patient-specific informations, to generate ECG signals representative of the target rare disease. This two-stage framework effectively integrates the knowledge representation capabilities of large language models with the generative capabilities of DiffuSETS.  By overcoming the data scarcity challenge, this approach enables the generation of realistic ECG signals for extremely rare cardiac conditions, providing a valuable resource for both educational and clinical applications. As an instance, We use this two-stage framework to generate ECG signal under the rare Brugada syndrome. The resulting ECG (Figure~\ref{fig: 4}B) shows clear signs of right bundle branch block and persistent ST segment elevation, both of which are the typical signs of Brugada syndrome and match the canonical waveform patterns depicted on cardiology literature\citep{brugada}. This result not only validates the proposed two-stage framework but also demonstrates that DiffuSETS is capable of generating ECG signals for extremely rare heart conditions.

Finally, we explore the potential of using DiffuSETS to initiate new medical discoveries, specifically focusing on uncovering hidden causal links between ECG signals and non-cardiac diseases. In previous training and experiments, we all utilized the ECG-related clinical text report as input, which contains abundant morphological information describing the ECG’s characteristics. Here, we demonstrate through two validated cases that DiffuSETS can also generate plausible ECG signals by taking input related to non-cardiac conditions. This capability highlights its potential to uncover new medical insights about the relationships between non-cardiac conditions and ECG signals.

In the first case, we show that DiffuSETS can generate ECG signals related to the patient’s heart age, showcasing its potential to aid in discovering age-related cardiological knowledge. Heart age\citep{heartage} is a clinically significant concept that reflects the functional and structural aging of the cardiovascular system, which is influenced by both intrinsic cardiac factors and external conditions, such as lifestyle, comorbidities, and vascular health. While heart age may not directly equal to chronological age, it also carries great medical significance, capturing the cumulative effects of aging and pathological processes on the heart and vasculature. To evaluate the ability of DiffuSETS to model heart age, we extract clinical text reports and patient-specific information from real ECG signals in the MIMIC-IV-ECG dataset, which are used as input for generating ECG signals. We then employ the AnyECG model\citep{anyECG} to estimate the heart age for both the real(reference) and generated ECG signals. The resulting scatter plot, shown in Figure~\ref{fig: 4}C, illustrates a significant correlation (r = 0.594, p < 0.001) between the heart ages of real(reference) and generated ECG signals. This finding demonstrates that DiffuSETS can effectively model age-related trends in ECG signals, highlighting its potential utility in cardiological research. For instance, it could play a key role in identifying age-specific ECG patterns, simulating rare cardiac conditions associated with particular age groups, and deepening our understanding of the relationship between aging and cardiovascular health.

In the other case, we examine hyperkalemia, a condition characterized by elevated potassium levels ($\mathrm{K}^+$) in the blood. While hyperkalemia is not a cardiac disease, it is well-known to impact ECG signals. To test if DiffuSETS can capture the latent relationship, we use the text input “The patient has hyperkalemia.” along with patient-specific information: “Gender: Male, Age: 50, Heart Rate: 90 bpm.” for generation. The resulting ECG signal (Figure~\ref{fig: 4}D) aligns with documented trends in ECG waveforms affected by rising potassium levels\citep{hyperkalemia}, including the presence of peaked T waves, widening of the QRS complex, and the progression of the ECG waveform toward a sinusoidal pattern. Even with alternative text input, such as “The patient has high potassium levels in the blood,” DiffuSETS consistently produces ECG signals exhibiting characteristic peaked T waves. This result underscores the model’s ability to identify and represent hidden causal relationships between non-cardiac conditions and ECG signals. Importantly, this capability is achieved despite the absence of explicit cardiac cues in the input conditions, potentially attributed to the similarities of deep embeddings derived from large language models, which capture associations between non-cardiac conditions and ECG-related clinical text reports. To some extent, DiffuSETS functions as an interpreter, visualizing the obscure medical knowledge embedded within large language models. Thus, whether as a method to validate hypotheses or a tool for inspiration, we believe DiffuSETS holds promise for uncovering new medical knowledge and expanding our understanding of the interactions between various medical conditions and physiological signals.

\section*{DISCUSSION}

The electrocardiogram (ECG) is an essential tool in cardiological diagnosis, serving as a foundation for detecting and understanding various heart conditions. However, as deep learning continues to revolutionize medical research, the lack of high-quality, annotated ECG data has become a critical bottleneck, especially when it comes to rare diseases where data is inherently limited. This scarcity hampers the development and evaluation of advanced models, reducing their effectiveness in real-world applications. While several studies have introduced generative models for synthesizing ECG signals, these models typically rely on small dataset, are conditioned solely on high-level labels, such as general disease categories, and often lack a systematic and robust methodology to assess the fidelity, diversity, and clinical relevance of the generated signals. More importantly, there is growing interest in exploring how these ECG generative models can have a broader impact on the medical domain. Beyond addressing data scarcity, such models hold significant promise for advancing cardiology education by providing diverse and condition-specific ECG signals for learning. Furthermore, they could contribute to medical knowledge discovery by enabling the exploration of other physiology conditions, paving the way for novel insights and innovations in healthcare.

Motivated by aforementioned concerns, we present DiffuSETS, a novel electrocardiogram (ECG) generative model that leverages clinical text reports and patient-specific information to produce ECG signals with high fidelity and strong semantic alignment. To validate our approach, we develop and test DiffuSETS on a comprehensive evaluation framework, which includes signal, feature and diagnostic levels, highlighting the fidelity and semantic accuracy of the generated ECG samples. DiffuSETS also outperforms baseline models, particularly in generating diverse ECG signals. Besides the objective numerical metrics, we conduct a Turing test, the results of which reveals that even experienced cardiologists struggle to distinguish between real and DiffuSETS-generated ECGs. For the aspect of application, DiffuSETS effectively addresses data scarcity by providing high-quality synthetic ECG signals, which significantly enhance downstream tasks. Finally, we demonstrate the potential of DiffuSETS in medical education and knowledge discovery, showcasing its broad applicability and value.

DiffuSETS can provide a comprehensive and adaptable solution to several pressing challenges in clinical research and healthcare. By tackling the issue of data imbalance, it enables the generation of synthetic ECG signals for rare and underrepresented conditions based on clinical text reports. This capability not only enriches dataset diversity but also enhances the robustness of machine learning models in real-world clinical applications, ensuring they perform effectively across a wide range of scenarios. Additionally, the ability of DiffuSETS to generate ECG signals that align closely with clinical reports creates new opportunities in medical education and research. Educators and learners can use these tailored ECGs to better understand the intricate relationships between symptoms, diseases, and their corresponding ECG patterns, offering a dynamic and practical way to study complex cardiac conditions. This feature also holds significant potential for advancing knowledge discovery, as it allows researchers to explore and uncover causal links and patterns within cardiovascular data that may not be immediately apparent in traditional datasets. Last but not least, it is important to note that DiffuSETS is a promising tool, and while we have highlighted only a few immediate applications, we believe its potential extends far beyond these examples, driven by the creativity and innovation of the research community.


Future directions for conditional ECG generation include enhancing DiffuSETS to facilitate the creation of digital twins by conditioning ECG generation on patient-specific ECG signals. This advancement would enable personalized healthcare predictions and solutions, offering individualized ECG data that could be instrumental in patient monitoring, providing a reliable basis for critical clinical decision-making. Another promising avenue is the development of a prospective ECG agent. By leveraging large language models, ECG generation can be conditioned on more diverse and aligned modalities such as clinical text, imaging data, or other diagnostic inputs. On top of that, integrating DiffuSETS into an ECG agent powered by large language models could further amplify the potential of agent. This combination could extend its capabilities beyond traditional applications, making it a versatile tool in fields such as cardiology education, real-time diagnosis, and the development of predictive healthcare systems. Together, these advancements would not only expand the applicability of ECG generative models but also drive innovation across a wider range of medical and research domains.

\subsection*{Conclusion}
In this paper, we introduce DiffuSETS, a 12-lead ECG generation model contioned on clinical text report and patient-specific information. DiffuSETS shows its supremacy in the three level evaluations and cardiologist Turing test. We also demonstrate and showcase the application of DiffuSETS for addressing data scarcity as well as promoting medical education and knowledge discovery.

\newpage

\section*{METHODS}


\subsection*{Related Works}
There are many studies currently attempting to address the generation of electrocardiogram (ECG) signals, but these methods have several limitations. Firstly, many models can only generate short-term time series \citep{yoon2019time, delaney2019synthesis, golany2020improving, li2022tts}, enabling them to produce only the content of a single heartbeat, rather than long-term ECG recordings. Secondly, they are often trained on small datasets with a limited number of patients \citep{zhu2019electrocardiogram}, or they use only a limited set of conditional labels \citep{golany2020simgans, sang2022generation, golany2019pgans}. In addition, many of these methods require ECG segmentation as a pre-training step, rather than directly processing noises \citep{golany2020simgans, sang2022generation, golany2020improving, li2022tts}. Moreover, many of these methods are capable of generating and classifying for specific patients only \citep{golany2019pgans}, and lack comprehensive training data and samples aimed at the general population \citep{thambawita2021deepfake}.

In recent studies, some researchers have attempted to apply diffusion models to the generation of ECGs \citep{adib2023synthetic}, treating ECGs as images rather than time series, and their methods were limited to the unconditional generation of single-lead ECGs. Moreover, from a quantitative performance evaluation perspective, these methods have not surpassed those based on GANs for generating ECGs. ME-GAN \citep{chen2022me} introduces a disease-aware generative adversarial network for multi-view ECG synthesis, focusing on how to appropriately inject cardiac disease information into the generation process and maintain the correct sequence between views. However, their approach does not consider text input, and therefore cannot incorporate information from clinical text reports. Auto-TTE \citep{chung2023text} proposed a conditional generative model that can produce ECGs from clinical text reports, but they also segmented the ECGs as a preprocessing step. SSSD-ECG \citep{alcaraz2023diffusion} introduced a conditional generative model of ECGs with a structured state space, encoding labels for $71$ diseases and incorporating them into the model training as conditions, but it cannot accept clinical text reports in the form of natural language text, thus lacking some of the rich semantic information inherent in disease diagnosis. At the same time, due to the lack of a unified performance evaluation setup, it is often challenging to quickly assess the relative merits of these methods.

\subsection*{Model Architecture}

The network architecture of DiffuSETS comprises a training phase and an inference phase, as depicted in Figure~\ref{fig: 1}B. \textbf{In the training phase,} we first extract 12-lead ECG signal $x$ from the ECG dataset. The signal-space representations of 12-lead ECG is then compressed by the encoder $E_\phi$ of variational autoencoder \citep{vae} to obtain latent-space representation of the ECGs \citep{rombach2022latent}, marked as $z_0$. Corresponding clinical text reports, after processing with prompts and utilizing an LLM, are transformed into a text embedding vector. Patient-specific information is also processed into a patient-specific embedding vector and merged with the text embedding vector to form a condition embedding vector $c$, which is then incorporated into the model's training. Subsequently, the denoising diffusion probabilistic model (DDPM, \citep{ho2020denoising}) scheduler continuously adds Gaussian noise $\epsilon_t$ to get the latent-space representation $z_t$ at randomly sampled time step $t$ through forward process formula: $z_t = \sqrt{\bar{\alpha}_t}z_{0} + \sqrt{1 - \bar{\alpha}_t}\epsilon_t\ ,\epsilon_t \sim \mathcal{N} (0, \mathbf{I})$ . 

The noise predictor, fed with the noisy latent-space representation $z_t$, current time step $t$ and the condition embedding vector $c$, is trained to predict that noise. The loss function of the training phase is defined as:
\begin{equation}
\label{eq: loss}
    \mathcal{L}_{\mathrm{DiffuSETS}} = \Vert \epsilon_t - \hat{\epsilon}_\theta(z_t, t, c) \Vert_2^2
\end{equation}
where $\hat{\epsilon}_\theta(z_t, t, c)$ stands for the output of noise prediction model. By performing gradient descend on Equation~\ref{eq: loss}, we can raise the Evidence Lower BOund (ELBO) so as to maximize the log likelihood of the training samples \citep{ho2020denoising}.

\textbf{In the inference phase,} the initial ECG signal latent $z_T$ is a noise vector sampled from the standard normal distribution. At each point during time step descends from $T$ to $1$, the noise prediction model attempts to predict a noise $\hat{\epsilon}_\theta(z_t, t, c)$ with the assistance of the input clinical text reports and patient-specific information. Then, the denoising diffusion probabilistic model scheduler denoises the latent-space representation $z_t$ to retrieve $z_{t-1}$ through a sampling process:  
\begin{align}
    &z _{t-1} \sim \mathcal{N} (\mu_q,\ \sigma_t^2 \mathbf{I})\\
    &\mu_q :=  
    [\sqrt{\alpha_t}(1-\bar{\alpha}_{t-1})z_t + 
    \sqrt{\bar{\alpha}_{t-1}}(1-\alpha_t) \hat{z}_0]
    / (1 - \bar{\alpha}_t) \\
    &\hat{z}_0 := [z_t - \sqrt{1-\bar{\alpha}_t}\hat{\epsilon}_\theta(z_t, t, c)]
    / \sqrt{\bar{\alpha}_t} \\
    &\sigma_t^2 := (1 - \alpha_t)(1 - \bar{\alpha}_{t-1}) / (1 - \bar{\alpha}_t)
\end{align}
where $\alpha_t$ is the hyperparameter related to diffusion forward process noise. Finally, our trained decoder $D_\theta$ reconstructs the normal 12-lead ECG signal based on the denoised latent-space representation, producing a signal-space ECG waveform series that aligns with the input descriptions.

\subsection*{Variational Autoencoder}

The variational autoencoder \citep{vae} consists of two parts: a encoder $E_\phi$ to compute the mean and variance of latent normal distribution of input ECG signal $x$ and a decoder $D_\theta$ to reconstruct the latent vector $z$ back to ECG signal. The latent-space representation is computed through reparameterization method (Equation~\ref{eq: repara}) to enable the gradient pass through the discrete sampling process. 
\begin{equation}
    \label{eq: repara}
    z \sim \mathcal{N}(\mu, \sigma^2) 
    \Longleftrightarrow\ z = \mu + \sigma \times \epsilon,\ \epsilon \sim \mathcal{N}(0, \mathbf{I})
\end{equation}
We train the variational autoencoder separately and whose loss function comprises two parts: reconstruction error and KL divergence. The reconstruction error uses Mean Squared Error (MSE) to measure the difference between the input ECG and the reconstructed ECG, while the KL divergence measures the difference between the encoded latent distribution and the standard normal distribution $N(0, 1)$. Combining these two parts, our loss function expression is: 
\begin{align}
    \mathcal{L}_{\mathrm{vae}} &= \mathrm{MSE}(x_{\mathrm{input}},\  x_{\mathrm{recons}}) + \lambda \cdot D_{KL} \left( q_\phi(z|x)\ \Vert\ \mathcal{N}(0, \mathbf{I}) \right) \\ 
    &= \frac1N\sum_{i = 1}^{N} (x_i - D_\theta(z_i))^2 - 
    \frac\lambda{2}\sum_{j = 1}^{N} (1 + \log(\sigma_j) - \mu_j^2 - \sigma_j^2)
\end{align}
where $q_\phi(z|x)$ is the latent-space variable distribution and $\mu_j$, $\sigma_j$ are the outputs of encoder $E_\phi$. In order to alleviate the KL vanishing problem \cite{bowman2015generating}, we adopt the monotonic KL-annealing where coefficient $\lambda$ starts at $0$ and increases linearly with the growth of epochs.

\subsection*{Noise Prediction Model}

\begin{figure}[!tbp]
  \centering
  \includegraphics[width=\linewidth]{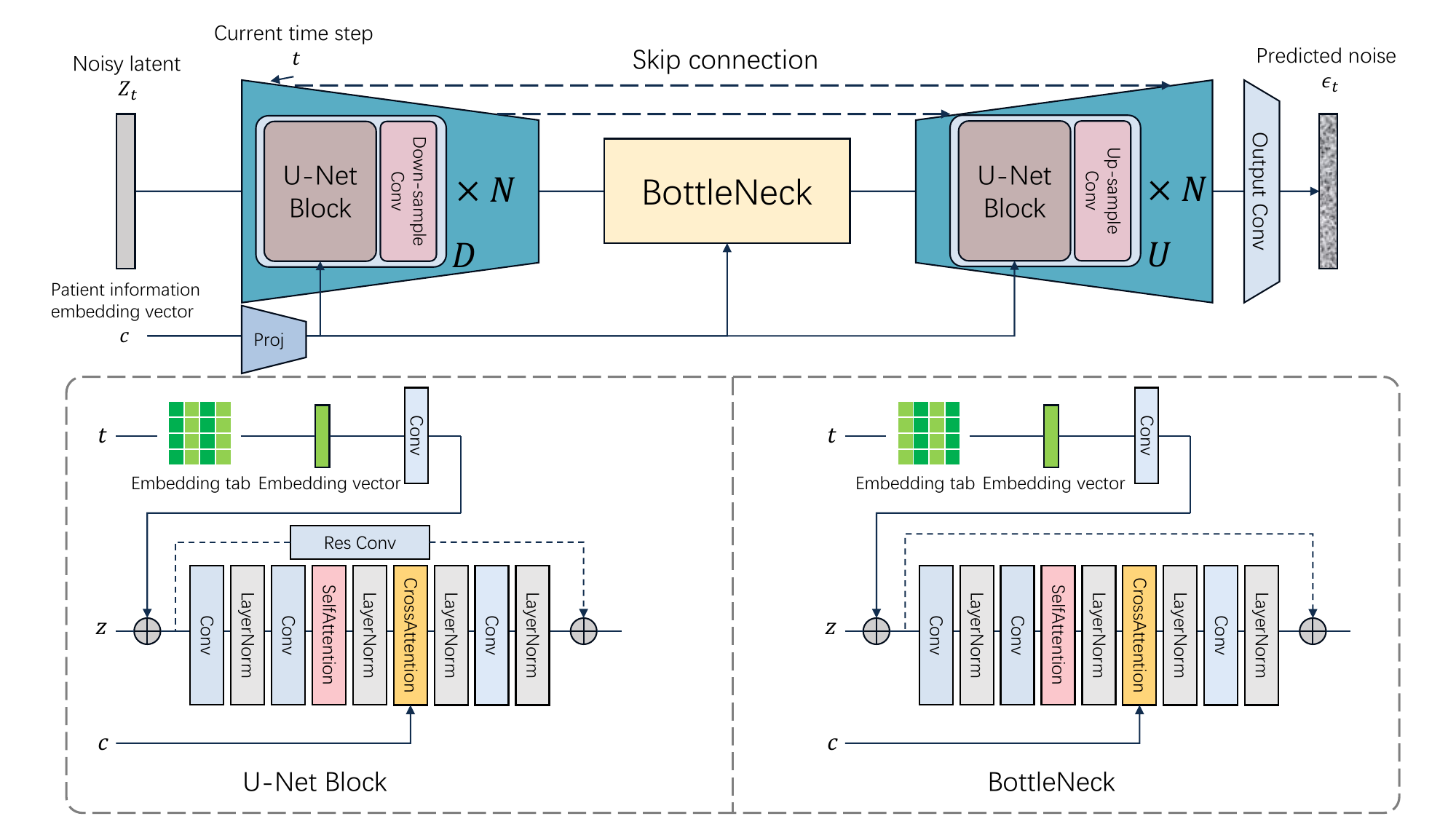}
  \caption{The detail architecture of noise predictor model in DiffuSETS.}
  \label{fig: noi pre}
\end{figure}

Our noise predictor follows the architecture of U-Net \citep{unet}, which contains a group of down-sampling layers $D_i$, a group of up-sampling layers $U_j$ and a bottleneck block concatenating two groups. The detailed architecture of noise predictor model are shown in Figure~\ref{fig: noi pre}. Passing through a down-sampling layer, the latent vector $z \in \mathbb{R}^{C \times L}$ would be enriched in channel dimension while be shortened in length dimension and vice versa. Besides the directly information flow from anterior layer $U_i$ to subsequent layer $U_{i+1}$ within the up-sampling groups, there also exist skip connections linking the down-sampling layer at the same level. Therefore the input expression of layer $U_i$ can be written as:
\begin{equation}
    In(U_i) = \mathrm{Concat}(Out(U_{i - 1}),\ Out(D_{i}))
\end{equation}
The noise prediction model takes three input: time step $t$, current latent-space representation $z_t$ and the condition embedding vector $c$. For time step $t$, we build a trainable embedding table to fetch time embedding and then add to $z_t$. The $t$-th row of time embedding table is initialized as:
\begin{equation}
    \mathrm{time\ emb} = \mathrm{Concat}
    \left( \left\{ \sin (t \cdot e^{-\frac{10i}{d/2 - 1}}) \right\}_{i=0}^{\frac{d}{2}-1},\ 
    \left\{ \cos (t \cdot e^{-\frac{10i}{d/2 - 1}}) 
    \right\}_{i=0}^{\frac{d}{2}-1}
    \right)
\end{equation}
where $d$ is the dimension of embedding length, and is assigned to $64$ in our model.

For condition embedding vector $c$, it is embraced in the cross attention block \citep{attention} in both sampling block and bottleneck block. Moreover, we deploy the self attention block \citep{attention} to consider the global details in latent vector, which promotes the consistency in QRS complex amplitude of generated ECG waveform.

\subsection*{Processing Clinical Text Reports}
\label{Processing Clinical Text Reports}
To achieve better semantic alignment with clinical text reports and patient-specific information, we design different processing methods of conditions based on the diverse data types and distributions. The results are then merged into an embedding vector to represent the patient's features. To enhance the model's ability to accept clinical text reports in natural language format as input, we also devised prompts for these texts and utilized the semantic embedding model "text-embedding-ada-002" provided by OpenAI (referred to as ada v2).

\begin{figure}[!tbp]
  \centering
  \includegraphics[width=0.9\linewidth]{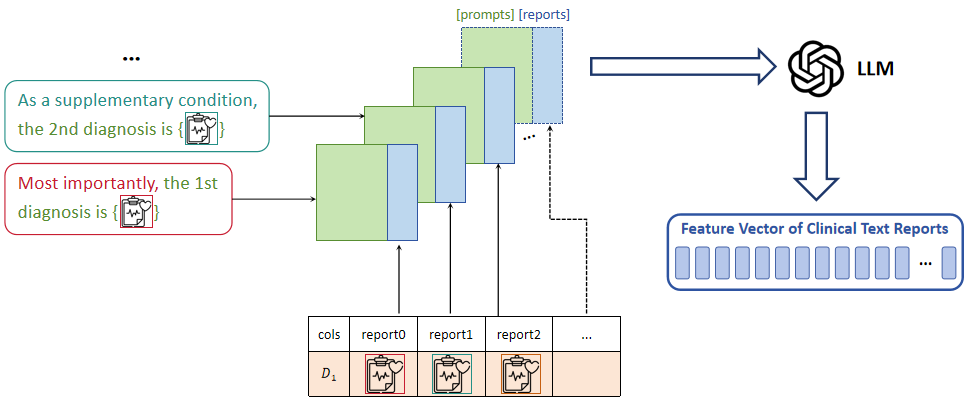}
  \caption{Processing of clinical text reports in DiffuSETS.}
  \label{clinical_reports}
\end{figure}

The processing workflow for clinical text reports in this paper is shown in Figure~\ref{clinical_reports}. We employed a pre-trained language model to process the clinical text reports. Specifically, for handling natural language text in clinical text reports, we use ada v2 to generate text embedding vectors. Before inputting the clinical text reports into ada v2, we designed prompts for processing. If only one report is inputted, the prompt is \textbf{"The report of the ECG is that \{text\}."} However, it is common for the dataset tables to show that one ECG corresponds to \textbf{multiple} clinical text reports, for which special arrangements have been made. In clinical datasets, the presence of multiple clinical text reports often serves to complement each other; typically, the most important report is placed first, with the remaining content supplementing the first report from various perspectives. Therefore, we designed specific ordered prompts for them. For the first clinical text report, our prompt is "Most importantly, The 1st diagnosis is \{text\}." For the subsequent reports, our prompt is "As a supplementary condition, the 2nd/3rd/... diagnosis is \{text\}." This enables the model to recognize the differences when processing multiple clinical text reports, thereby better understanding the semantic information contained within the reports.

\subsection*{Processing Patient-Specific Information}
In the MIMIC-IV-ECG and MIMIC-IV-Clinical datasets, there is a wealth of tabular data recording patient-specific information. We categorize these characteristics into three types: categorical demographic condition, numerical demographic condition, and other health condition. We have designed specific processing methods for each type of data, consolidating their information into a patient-specific embedding. 

\textit{Categorical Demographic Condition}. We categorize discrete, categorical conditions such as sex and race as categorical demographic conditions. In our DiffuSETS method, we use the feature of sex to represent categorical demographic conditions in experiments. Since sex is binary data, it can be represented simply using $0$ or $1$.

\textit{Numerical Demographic Condition}. We categorize continuous, numerical conditions such as age and weight as numerical demographic conditions. These types of data are stored in tables in numerical form. During the training and inference processes of the model, they can be directly utilized. In our DiffuSETS method, we use the feature of age to represent numerical demographic conditions in experiments. For this category of conditions, it is important to consider the data distribution and the removal of outliers.

\textit{Other Health Condition}. Specifically, we categorize data related to patient health metrics such as heart rate and left ventricular ejection fraction (LVEF) as other health conditions. They can also affect the morphology of the ECG. Many of these types of data are recorded in dataset tables, and others require processing to be obtained. Notably, when an ECG is provided, these values can often be calculated. Therefore, in the task of generating ECGs, we can perform calculations on the generated ECGs to intuitively assess the generation effectiveness. In our DiffuSETS method, we use heart rate to represent other health conditions in experiments and have conducted feature-level evaluation and analysis of this characteristic after generating the ECG, making full use of the data's intrinsic properties.

\begin{figure}[!tbp]
  \centering
  \includegraphics[width=1.0\linewidth]{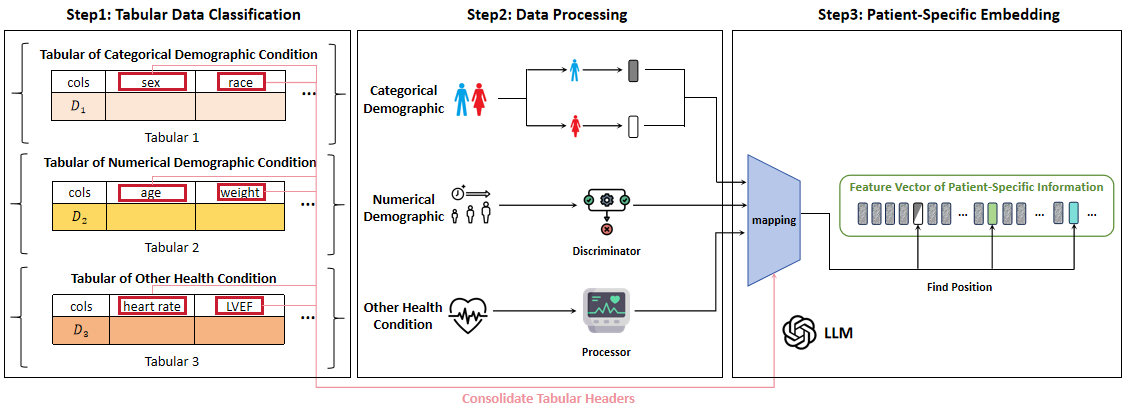}
  \caption{Processing of patient-specific information in DiffuSETS.}
  \label{patient_specific}
\end{figure}

Finally, we concatenate the processed patient-specific embedding with the text embedding vector generated from the clinical text reports to obtain the conditions embedding vector $c$ (Using sex, age, and heart rate as examples, see Equation~\ref{eq: ps info}). This vector is used for both model training and inference, facilitating the model's understanding of the semantic information included in the input. The above content is illustrated in the Figure~\ref{patient_specific}.
\begin{equation}
    \label{eq: ps info}
    \resizebox{0.9\columnwidth}{!}{$
    c = \mathrm{Concat}\left( \mathrm{ada\_v2}(text),\ hr,\ age,\ G(sex) \right),\ 
    G(x) = \left\{ \begin{array}{ll} 0, & x = \mathrm{F} \\ 1, & x = \mathrm{M} \end{array}\right.
    $}
\end{equation}

\subsection*{Datasets}
We use the MIMIC-IV-ECG dataset \citep{gowmimic} to train the DiffuSETS model. MIMIC-IV project covers hospital admission records of $299,712$ patients from 2008 to 2019 at the Beth Israel Deaconess Medical Center, including patient personal information such as age and sex. The ECG dataset within it contains $800,035$ records with ECG signals, patient IDs, RR intervals, and machine-generated clinical text reports. For each signal, we search the sex and age characteristics of the ECG owners in MIMIC-IV-Clinical\citep{johnson2023mimic} patient table by the patient IDs, and calculate the heart rate using the RR intervals. However, some RR intervals showed anomalies, such as $0$ ms or $65,535$ ms. Therefore, for data samples where the RR intervals fall outside the range of $300$ ms to $1,500$ ms, we use the XQRS detector from the wfdb toolkit \citep{sharma2023wfdb} to obtain the QRS intervals through waveform analysis to calculate the heart rate. Samples that could not calculate a heart rate from all 12 waveforms are considered to have abnormal heart rate records and are discarded along with samples missing sex or age information. After the preprocessing, we retain $794, 372$ records. Each lead's original data is a 10-second ECG signal at a sampling rate of $500$ Hz, resulting in $5,000$ time samples. We down-sample these to $1,024$ time samples for model training and internal validation. We also use the PTB-XL dataset for external validation, which contains $21,799$ clinical entries, each with a 10-second ECG signal, along with patient-specific information and cardiologist-recorded ECG reports. The PTB-XL labels do not include records of heart rate, so we directly use the waveforms to calculate the heart rate. Similarly, we sample the ECGs at a rate of $500$ Hz and down-sample them to $1,024$ time samples, following the processing method used for MIMIC-IV-ECG dataset.

\subsection*{Implementation Details}

Our method is trained on a GeForce RTX 3090 using PyTorch 2.1. Batch size is set to $512$, with a learning rate of $5 \times 10^{-4}$. The latent space is set to $\mathbb{R}^{4 \times 128}$. The number of time step $T$ in training phase is set to $1,000$ while noise $\beta_{t}$ of diffusion forward process are assigned to linear intervals of $[0.00085, 0.0120]$. Noise predictor has $7$ layers and the kernel size of convolution is $7$. It iterates approximate 60 time steps per second within the same environment in inference phase.

\subsection*{Metrics}

The signal level distribution similarity is evaluated by the Fréchet Inception Distance (FID) score (the lower is better). It is a widely used metric for evaluating the quality of generated data, particularly in generative models. Originally developed for assessing image generation, the FID score measures the similarity between the distributions of real and generated data. It computes the Fréchet distance (or Wasserstein-2 distance) between feature embeddings extracted from a pre-trained model. $\mu_r$ and $\Sigma_r$ are the mean and covariance of the real data’s feature embeddings, and $\mu_g$ and $\Sigma_g$ are those of the generated data.

\begin{equation}
    \mathrm{FID}=\|\mu_{r}-\mu_{g}\|^{2}+\mathrm{Tr}(\Sigma_{r}+\Sigma_{g}-2(\Sigma_{r}\Sigma_{g})^{1/2}),
\end{equation}

The precision, recall, and F1 score (all of which are higher when better) for generative models are computed by evaluating the overlap between the representation points and representation manifolds\cite{kynkaanniemi2019improved}. A representation manifold is defined as union of hyperspheres centering at each representation point, where the radius is distance to the k-th nearest neighbourhood ($k=3$ in our experiments). Precision is determined by the proportion of generated points $g$ that fall within the real signal manifold, while recall measures the proportion of real signal points $x$ covered by the generated manifold. The F1 score, as the harmonic mean of precision and recall, provides a balanced assessment of the model’s ability to produce accurate and diverse outputs. Conceptual cases about the calculation of precision and recall are depicted in Figure~\ref{fig: 2}C.

\begin{align}
    &f(e,\{E\})=\left\{\begin{array}{ll}1,& \exists\ e' \in \{E\}\ s.t. \left\|e-e'\right\|_2\leq\left\|e'-\text{NN}_k\left(e',\{E\}\right)\right\|_2\\
    0,& \text{otherwise}\end{array}\right. 
\end{align}
\begin{align}
    \text{Precision}(\{X\},\{G\})&=\frac{1}{|\{G\}|}\sum_{g\in\{G\}}f(g,\{X\}) \\
    \text{Recall}(\{X\},\{G\})&=\frac{1}{|\{X\}|}\sum_{x\in\{X\}}f(x,\{G\})\\
    \text{F1}(\{X\},\{G\})&=2\cdot\frac{\text{Precision}\cdot\text{Recall}}{\text{Precision}+\text{Recall}}
\end{align} 

In feature level, the Heart Rate MAE (the smaller is better) are calculated as the mean of absolute error between heart rate ($\mathrm{hr_{ref}}$) in input patient specific information and heart rate ($\mathrm{hr_{gen}}$) derived from generated ECG signals.

\begin{equation}
    \text{Heart Rate MAE} = \frac1n \sum_{i=1}^n |\mathrm{hr_{ref}}^i - \mathrm{hr_{gen}}^i|
\end{equation}

In diagnostic level, we use the classic CLIP score\citep{clipscore}, which quantifies how well a generated sample matches a given text description based on the similarity of their representations in the shared latent space of the Contrastive Language–Image Pretraining (CLIP) model. In our experiments, the text (in form of LLM embedding) and ECG signal inputs are first encoded using the respective text and signal encoders of the CLIP model. The cosine similarity between the resulting representations $R_\text{text}$ and $R_\text{signal}$ is then calculated as CLIP score.

\begin{equation}
    \text{CLIP Score} = \frac{R_\text{text} \cdot R_\text{signal}}{\|R_\text{text}\| \|R_\text{signal}\|}
\end{equation}

\newpage


\section*{RESOURCE AVAILABILITY}

The code of our method and evaluations is publicly available at \url{https://github.com/Raiiyf/DiffuSETS_Exp}.

\noindent All of the datasets in our work is pubblicly available: MIMIC-IV-ECG, \url{https://physionet.org/content/mimic-iv-ecg/1.0/}; MIMIC-IV-Clinical, \url{https://physionet.org/content/mimiciv/3.1/}; PTB-XL, \url{https://physionet.org/content/ptb-xl/1.0.3/}.

\section*{ACKNOWLEDGMENTS}


This work was supported by the National Natural Science Foundation of China (62102008, 62172018); CCF-Zhipu Large Model Innovation Fund (CCF-Zhipu202414).




\section*{DECLARATION OF INTERESTS}


The authors declare no competing interests.

\newpage


\bibliography{reference}

\bigskip


\newpage

\end{document}